\documentclass{article}

\usepackage[preprint]{neurips_2026}

\usepackage{graphicx}
\usepackage{amsmath}
\usepackage{algorithm}
\usepackage{algpseudocode}
\usepackage{wrapfig}

\usepackage[utf8]{inputenc} 
\usepackage[T1]{fontenc}    
\usepackage{hyperref}       
\usepackage{url}            
\usepackage{booktabs}       
\usepackage{amsfonts}       
\usepackage{nicefrac}       
\usepackage{microtype}      
\usepackage{xcolor}         
\usepackage{placeins}

\usepackage[utf8]{inputenc}
\usepackage[T1]{fontenc}
\usepackage{microtype}
 
\usepackage{amsmath, amssymb, amsthm, amsfonts}
\usepackage{bm}
\usepackage{mathtools}
 
\usepackage{booktabs}
\usepackage{multirow}
\usepackage{array}
\usepackage{graphicx}
\usepackage{caption}
\usepackage{subcaption}
\usepackage{xcolor}
\usepackage{pifont}            
 
\usepackage{algorithm}
\usepackage{algpseudocode}
\newtheorem{proposition}{Proposition}
\newtheorem{remark}{Remark}

\newcommand{\method}{\textsc{AdaMerge}}
\newcommand{\tome}{\textsc{ToMe}}
\newcommand{\R}{\mathbb{R}}

\newcommand{\cmark}{\ding{51}}
\newcommand{\xmark}{\ding{55}}

\definecolor{daeuncolor}{RGB}{0,102,204}

\title{AdaMerge: Salience-Aware Adaptive Token Merging\\
for Training-Free Acceleration of Vision Transformers}

\author{%
  Semi Lee \\
  Electronic Engineering\\
  Soongsil University\\
  Seoul, South Korea \\
  \texttt{semi2223@soongsil.ac.kr} \\
  \And
  Hyejin Go \\
  Electronic Engineering\\
  Soongsil University\\
  Seoul, South Korea \\
  \texttt{hyejin1612@soongsil.ac.kr} \\
  \And
  Hyesong Choi\thanks{Corresponding author.} \\
  Electronic Engineering\\
  Soongsil University\\
  Seoul, South Korea \\
  \texttt{hyesong@ssu.ac.kr} \\
}

\begin{document}

\maketitle

\begin{abstract}
The quadratic cost of self-attention in Vision Transformers (ViTs) constitutes 
a fundamental bottleneck for practical deployment, motivating a vibrant line of 
research on token reduction. Among existing approaches, token merging 
(\tome{}) has emerged as an elegant training-free solution; yet its design rests 
on an unspoken premise---token equality---which contravenes the 
well-documented non-uniformity of self-attention and leads to \textbf{information 
loss} in high-salience tokens under aggressive compression. We address 
this limitation with \textbf{\method{}}, a token-merging framework based 
on two complementary mechanisms. First, \emph{salience-weighted similarity} 
leverages column-wise feature-affinity centrality as a token-importance proxy and incorporates 
the resulting salience scores into the bipartite matching score, ensuring that informationally pivotal 
tokens contribute more strongly to the merged representation. Second, \emph{adaptive merging 
intensity} uses pre-computed layer-wise similarity statistics 
($\mu_l, \sigma_l$) to dynamically modulate the per-layer reduction count $r_l$ 
in accordance with input-specific redundancy. On ImageNet-1k validation with 
ViT-B/16, \method{} \textbf{consistently outperforms \tome{}}, PiToMe, and DSM across 
all FLOPs-matched regimes. The accuracy gap widens 
monotonically with compression intensity: at the extreme ${\sim}13.4$G FLOPs 
operating point, \method{} sustains a Top-1 accuracy degradation of only 
$-1.06\%$, compared to $-1.45\%$ for PiToMe and $-4.62\%$ for DSM --- 
indicating that fixed merging schedules can degrade substantially under aggressive compression. 
This monotonic widening is consistent with the hypothesis that uniform and schedule-rigid merging 
can incur increasing information loss as compression grows. \textbf{To our knowledge, \method{} is the first to combine} 
salience-weighted similarity \emph{and} adaptive per-layer reduction into a 
single training-free token merging framework, advancing the accuracy--FLOPs 
Pareto frontier of ViT acceleration. Code is available in the Supplementary Material.
\end{abstract}
\section{Introduction}
\label{sec:intro}

Vision Transformers~\citep{dosovitskiy2021vit} have outperformed
convolutional architectures on image-recognition benchmarks, yet their
$\mathcal{O}(N^{2}d)$ self-attention complexity poses a formidable
obstacle to real-time deployment. Within this context, \emph{token
reduction} has crystallized as a primary acceleration paradigm. The
literature largely bifurcates into \textbf{token pruning}
methods~\citep{rao2021dynamicvit,yin2022avit}, which discard tokens
outright at the cost of irrevocable information loss, and \textbf{token
merging} approaches, exemplified by \tome{}~\citep{bolya2023tome}.
\tome{} consolidates similar tokens, attenuating information loss while
affording the singular practical virtue of operating in a
\emph{training-free} regime.

Despite \tome's empirical success, its design embodies a consequential
implicit premise: \textbf{token equality}. Its bipartite soft-matching
procedure selects pairs solely based on cosine similarity and aggregates
them via uniform averaging. This assumption stands in marked tension
with the well-documented non-uniform character of ViT
attention~\citep{caron2021dino,chefer2021transformer}---certain tokens
shoulder a disproportionate share of discriminative reasoning, while
others represent background redundancy. We contend that treating all
tokens equally causes an asymmetric erosion of high-salience
information, a penalty that compounds nonlinearly under aggressive
compression.

To confront this, we introduce \method{}, a framework that reconceives
token merging along two orthogonal axes: \emph{which} tokens to merge
and \emph{how aggressively} to merge them.

\textbf{Salience-Weighted Similarity.} We modulate the matching score
using a feature-affinity-based salience proxy. Through a
salience-proportional aggregation rule, we ensure that high-salience
features contribute more strongly to the merged representation.

\textbf{Adaptive $r$ via Layer-wise Statistics.} Moving beyond rigid
reduction schedules, we dynamically modulate the per-layer reduction
count $r_l$ based on pre-computed layer-wise similarity statistics and
input-specific redundancy. This affords \emph{dual adaptivity}---both
input-level and layer-level---not achieved in prior training-free
methods.

Our contributions are threefold.
\begin{itemize}
    \item \textbf{(C1) Salience-aware merging under a
    feature-reconstruction objective.} We recast \tome{} as the special
    case of uniform salience and formally show that salience-weighted
    matching yields a non-negative reduction in reconstruction error
    relative to uniform averaging.

    \item \textbf{(C2) Dual-adaptive merging-intensity mechanism.}
    To our knowledge, the first training-free token-reduction scheme
    that jointly achieves importance-aware matching \emph{and}
    input- and layer-adaptive compression intensity, supported by an
    iterative statistics-refinement protocol.

    \item \textbf{(C3) Pareto Frontier Extension.} On ImageNet-1k
    with ViT-B/16, \method{} consistently outperforms \tome{}, PiToMe,
    and DSM at matched FLOPs. The accuracy gap widens monotonically
    with compression intensity, sustaining only $-1.06\%$ degradation
    at the extreme ${\sim}13.4$G FLOPs tier versus $-4.62\%$ for DSM,
    supporting the value of adaptive merging over rigid schedules.
\end{itemize}
\section{Related Works}
\label{sec:related}

\subsection{Token Pruning}
Token pruning reduces computation by selecting and discarding less
informative tokens. DynamicViT~\citep{rao2021dynamicvit} introduces a
learned prediction module to emit per-token retention probabilities,
while A-ViT~\citep{yin2022avit} utilizes Adaptive Computation Time for
token-level early exits. EViT~\citep{liang2022evit} leverages
[CLS]-token attention as an importance prior. Subsequent works explore
interpretability-aware redundancy reduction, adaptive token sampling,
patch slimming, and explicit ViT
pruning~\citep{pan2021iared2,fayyaz2022ats,tang2022patchslimming,zhu2021vitpruning}.
Although effective, these methods share two primary limitations:
(i)~discarded tokens incur permanent information loss, and
(ii)~many require costly fine-tuning or auxiliary training.

\subsection{Token Merging and Pooling}
Token merging preserves more information than pruning by consolidating
tokens rather than discarding them. \tome{}~\citep{bolya2023tome}
pioneered a selective bipartite soft-matching scheme that operates
without retraining, providing a foundation for plug-and-play
acceleration. Subsequent training-free refinements include
ToFu~\citep{kim2024tofu}, which explores hybrid pruning-merging
strategies, DSM~\citep{heo2024dsm}, which defers merging to deeper
layers to preserve local features, and PiToMe~\citep{tran2024pitome},
which utilizes an energy-based metric to protect informative tokens
from being merged. Input-adaptive vision methods such as
AdaViT~\citep{meng2022adavit}, Evo-ViT~\citep{xu2022evovit}, and
SPViT~\citep{kong2022spvit} perform importance-aware token selection
but require additional training. In contrast, \method{} jointly
addresses \emph{which} tokens to merge and \emph{how many} in a
strictly training-free regime---combining an importance-aware matching
signal with input- and layer-adaptive compression intensity.
\section{Motivation: Why Salience Matters}
\label{sec:motivation}

\subsection{Observation 1: Tokens Are Not Informationally Equal}
The non-uniform character of self-attention is by now well established.
DINO~\citep{caron2021dino} elucidated that the [CLS] token attends
preferentially to object regions, intimating an informational hierarchy
among patch tokens. Subsequent interpretability
analyses~\citep{chefer2021transformer, raghu2021vision, park2022how}
corroborate that a small subset of tokens commands a disproportionate
share of information flow, while the majority encode redundant background
content---a pattern that manifests as attention artifacts in
low-informative regions~\citep{darcet2023registers}.
Were patches genuinely interchangeable, attention maps would approach
uniformity; in practice, salience concentrates sharply on semantically
discriminative regions.

\begin{figure}[t]
\centering
\includegraphics[width=\linewidth]{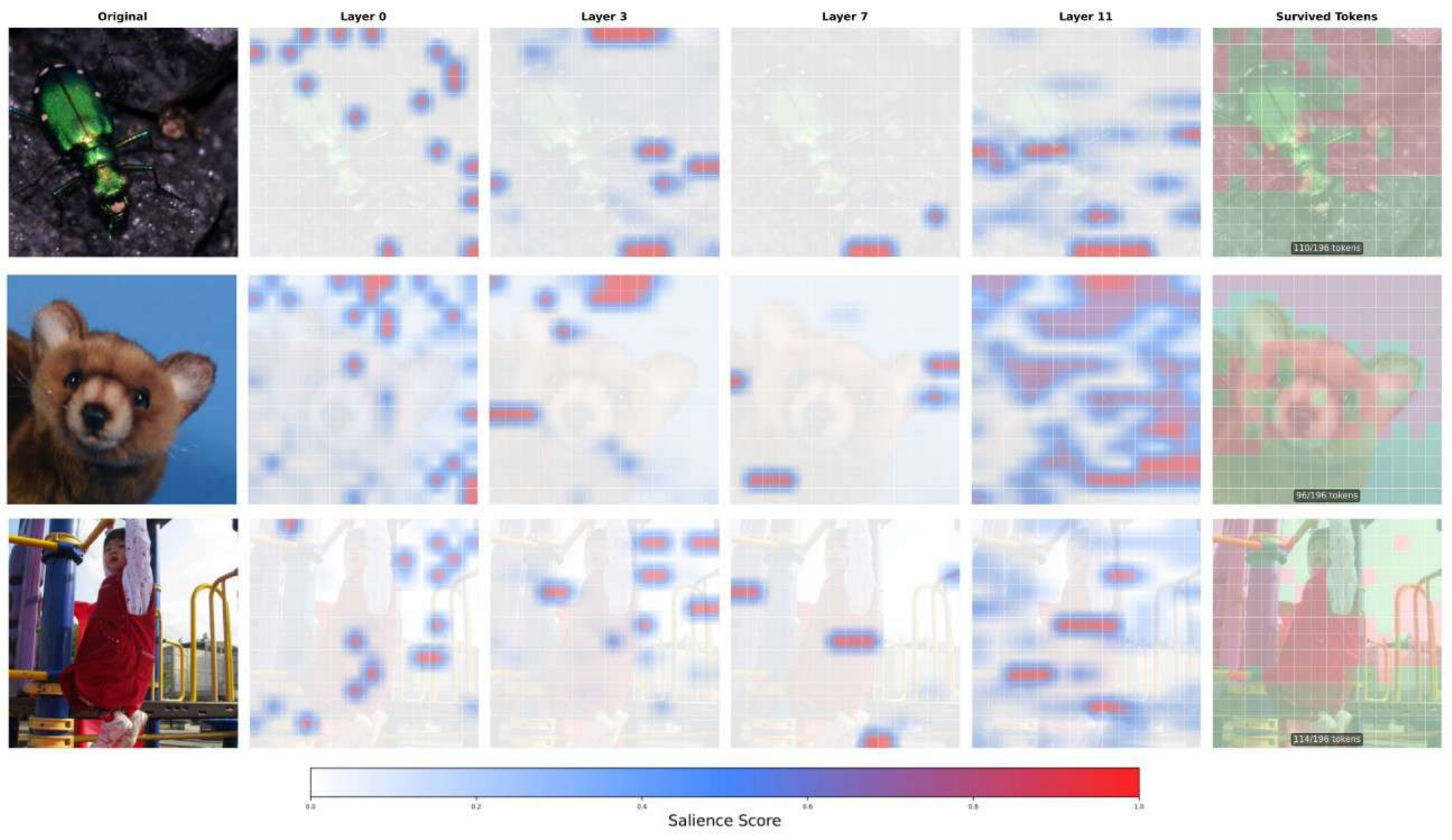}
\caption{
    Layer-wise salience maps and survived tokens from \method{}($r_{\max}{=}18$) on three ImageNet-1k images. Warmer colors indicate higher salience (column-wise sum of the row-normalized affinity matrix). Salience consistently localizes to discriminative regions and sharpens with depth, corroborating the token-importance non-uniformity motivating our design. In the rightmost column, green denotes survived tokens and red denotes merged tokens; object-centric patches are preferentially preserved. \textbf{Crucially, the varying proportion of merged tokens across different images demonstrates how \method{} content-adaptively allocates its merging budget to preserve semantic integrity.}
}
\label{fig:salience_vis}
\end{figure}

Figure~\ref{fig:salience_vis} provides direct empirical corroboration. The per-token salience scores consistently localize to semantically meaningful regions rather than background patches, and this localization sharpens progressively with depth. Crucially, the rightmost column demonstrates that \method{}'s salience-guided merging preserves object-centric tokens (green) while absorbing redundant background tokens (red), suggesting that our salience measure provides a useful proxy for token importance.

\subsection{Observation 2: Layer-wise Redundancy Is Heterogeneous} 
A second axis of non-uniformity manifests across network depth. Shallow layers process local texture and exhibit substantial redundancy among neighbouring patches; merging here carries little semantic risk. Deeper layers, by contrast, perform semantic abstraction where token representations diverge substantially, each encoding a distinct aspect of the global scene. A fixed merging count $r$ applied uniformly across all layers is therefore a poor fit: it over-merges in semantically rich deep layers and under-merges in redundant shallow ones. Capturing this heterogeneity quantitatively motivates our layer-wise statistics $(\mu_l, \sigma_l)$.

\subsection{The Failure of Existing Token Merging Methods}
These two observations expose critical failure modes shared by uniform merging methods like \tome~\citep{bolya2023tome}.

\paragraph{Merge-Without-Priority.}
When a high-salience token and a background token are consolidated via uniform averaging, the salient information is immediately halved. High-salience features are asymmetrically eroded. Furthermore, uniform matching procedures are oblivious to salience, leaving high-salience tokens equally vulnerable to absorption into less informative counterparts. While PiToMe~\citep{tran2024pitome} explores energy-based partitioning, its reliance on uniform averaging still dilutes critical features upon merging.

\paragraph{One-Size-Fits-All.}
\tome{}, PiToMe, and DSM~\citep{heo2024dsm} enforce fixed reduction counts regardless of input complexity or layer-wise redundancy. Complex images with high token diversity are over-compressed, incurring semantic blurring, while simple images waste capacity on unnecessary merges. DSM's \texttt{delay\_layer} heuristic, while deferring early merges, remains agnostic to per-image redundancy and collapses under extreme compression.

\paragraph{Failure amplifies with compression.}
As $r$ increases, the pool of candidate merging pairs inevitably
expands to include progressively more dissimilar, semantically
heterogeneous tokens. When $r$ is small, the damage of uniform
averaging remains contained. As $r$ grows, the asymmetric erosion
of salient tokens compounds super-linearly. This yields a concrete
prediction: \emph{the accuracy advantage of salience-aware merging
over uniform merging should widen monotonically as compression
intensifies}---a prediction our experiments support
(\S\ref{sec:main_results}).

\subsection{Design Principles}
The preceding observations crystallize into two design principles for \method{}.
\begin{itemize}
    \item \textbf{(P1) Survivor selection should be salience-biased.}
    The token that endures merging ought to inherit the richer
    informational content, ensuring that high-salience features
    contribute more strongly to the merged representation.

    \item \textbf{(P2) Merging intensity should be input- and
    layer-adaptive.} Compression should be aggressive where redundancy
    is high and conservative where token diversity prevails, adapting
    to both the input image and the current layer's semantic density.
\end{itemize}
\section{Method: \method}
\label{sec:method}

\begin{figure}[t]
\centering
\includegraphics[width=\linewidth]{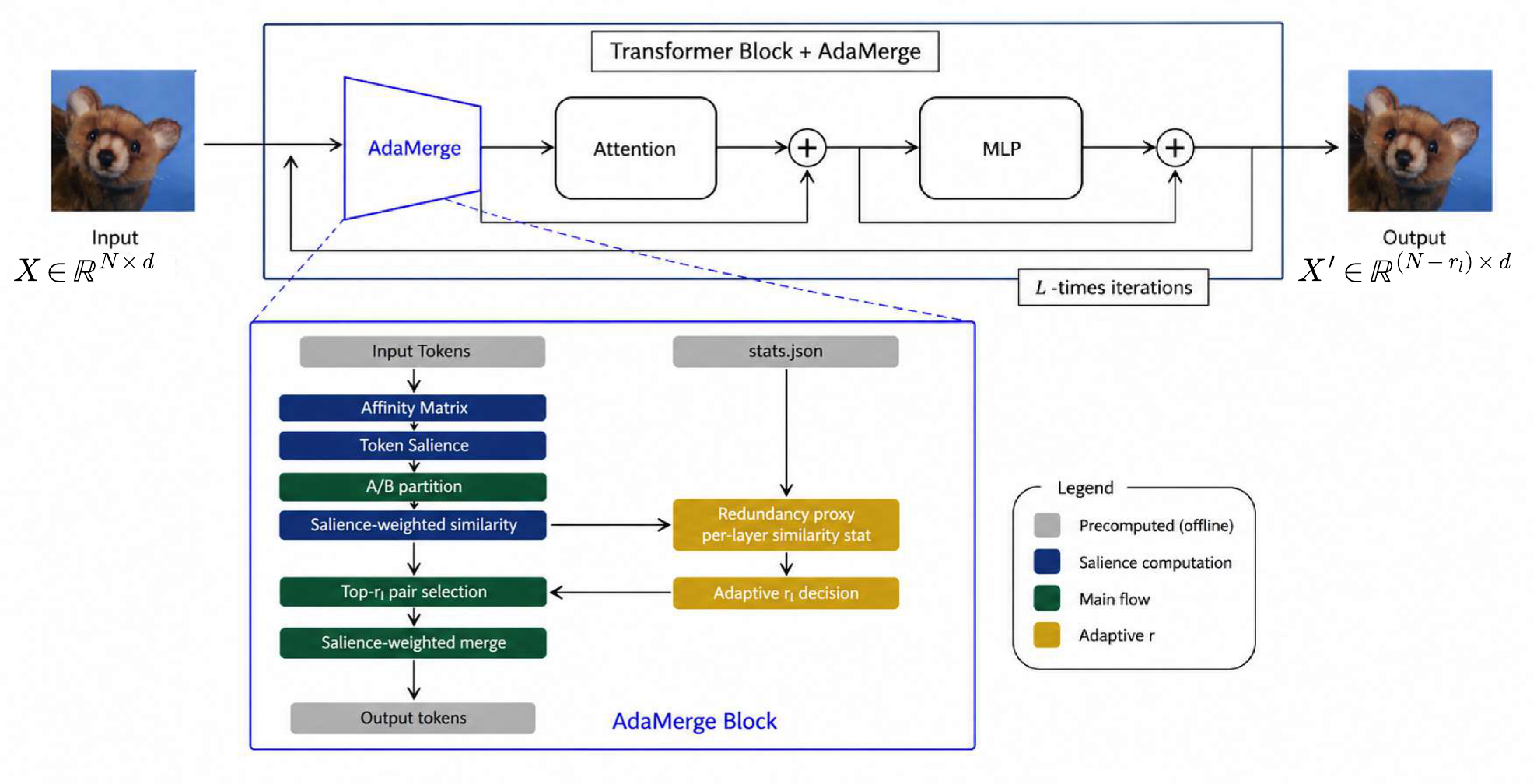}
\caption{
    \textbf{Top:} \method{} is integrated before each Transformer block to reduce the sequence length from $N$ to $N - r_l$ prior to self-attention.
    \textbf{Bottom:} Internal pipeline. Salience computation and adaptive $r_l$ decision run in parallel, followed by salience-proportional aggregation. Layer-wise statistics $(\mu_l, \sigma_l)$ are precomputed offline and loaded from \texttt{stats.json}.
}
\label{fig:method_overview}
\end{figure}
 
\subsection{Preliminaries and Notation}
Let $\mathbf{X} = [x_1, \ldots, x_N] \in \R^{N \times d}$ denote the patch-token sequence within a ViT block ($N=196$ for ViT-B/16). The CLS token is excluded from all merging operations and reattached to the reduced patch-token sequence after each merge step. We partition $\mathbf{X}$ into two equal-sized sets by sequential index split: $\mathcal{A} = \mathbf{X}_{[:N/2]}$ and $\mathcal{B} = \mathbf{X}_{[N/2:]}$. Unlike \tome~\citep{bolya2023tome}, which uses index-parity partitioning, this sequential split simplifies vectorized implementation and avoids spatial interleaving artifacts. The affinity matrix, normalized row-wise via softmax, is defined as:
\begin{equation}
    A_{ij} = x_i^\top x_j, \qquad \hat{\mathbf{A}} = \mathrm{softmax}(\mathbf{A}).
\end{equation}
 
\subsection{Salience-Weighted Similarity}
\label{subsec:salience_sim}
 
\paragraph{Token Salience.}
Following the feature-affinity salience formulation of SBAM~\citep{choi2024sbam}, we quantify token importance as feature-affinity centrality, defined as the column-wise sum of the row-normalized affinity matrix:
\begin{equation}
    s_i = \sum_{j=1}^{N} \hat{A}_{ji}, \qquad \mathbf{s} = \hat{\mathbf{A}}^\top \mathbf{1}.
\end{equation}
To ensure stability across layers, scores are min-max normalized to $[0, 1]$, yielding $\hat{s}_i$. 
This formulation requires no auxiliary pass---the affinity matrix $\hat{\mathbf{A}}$ is already computed during merging---and furnishes a per-layer signal that adapts to the evolving token distribution across depth.
 
\paragraph{Weighted Similarity Score.}
Let $\mathbf{s}^{\mathcal{A}}$ be the salience restricted to $\mathcal{A}$. We define the salience-weighted similarity as:
\begin{equation}
    \tilde{S}^{\mathcal{AB}}_{ij} = s^{\mathcal{A}}_i \cdot \cos(x^{\mathcal{A}}_i, x^{\mathcal{B}}_j), \qquad i \in \mathcal{A},\; j \in \mathcal{B}.
\end{equation}
The matching procedure identifies $\mathrm{best\_match}(i) = \arg\max_j \tilde{S}^{\mathcal{AB}}_{ij}$, prioritizing high-salience tokens for merging.
 
\subsection{Adaptive $r$ via Layer-wise Similarity Statistics}
\label{subsec:adaptive_r_method}

To capture heterogeneity across layers, we define a redundancy
proxy $\bar{S}_l$ as the mean of the maximum weighted similarities
at layer $l$. We precompute $(\mu_l, \sigma_l)$ using a 1\% subset
of ImageNet-1k. To ensure self-consistency, we employ an iterative
refinement protocol, which typically
converges within three iterations.

At inference, the merging count $r_l$ for an input is determined by
its layer-wise $z$-score, modulating compression intensity per layer
in the spirit of input-adaptive computation~\citep{graves2016act}:
\begin{equation}
    z_l = \frac{\bar{S}_l - \mu_l}{\sigma_l}, \qquad
    r_l = \lfloor r_{\max} \cdot \sigma(\alpha \cdot z_l) \rfloor.
\end{equation}
The sigmoid temperature $T$ (default $T{=}1.0$) controls merging
sharpness via $z_l \leftarrow z_l / T$; $T{=}0.5$ yields higher
accuracy at the cost of a sharper boundary.
 
\subsection{Bipartite Merge with Salience Aggregation}
For each selected pair $(i, j) \in \mathcal{M}$, we aggregate features via salience-proportional weighting to ensure informational integrity:
\begin{equation}
    \tilde{x} = \frac{s^{\mathcal{A}}_i \cdot x^{\mathcal{A}}_i + s^{\mathcal{B}}_j \cdot x^{\mathcal{B}}_j}{s^{\mathcal{A}}_i + s^{\mathcal{B}}_j}.
\end{equation}
The salience of the merged token is set to preserve the stronger signal:
\begin{equation}
    \hat{s}_{\tilde{x}} = \max\!\left(s^{\mathcal{A}}_i,\, s^{\mathcal{B}}_j\right).
\end{equation}
This ensures that high-salience features have greater weight in the resulting sequence.

\begin{proposition}[Salience-Weighted Aggregation Reduces Reconstruction Error]
\label{prop:reconstruction}
Under the noise model ($\eta_i \sim \mathcal{N}(0, \tau^2 I)$), for any matched pair $(i, j)$ with saliences $s_i, s_j > 0$, the salience-proportional aggregation $\tilde{x} = (s_i x_i + s_j x_j)/(s_i + s_j)$ satisfies
\begin{equation}
    \ell_{\mathrm{uniform}}(i,j) - \ell_{\mathrm{AdaMerge}}(i,j)
    \;=\;
    \frac{(s_i - s_j)^{2}}{2(s_i + s_j)^{2}}\,\|x_i - x_j\|^{2}
    + \mathcal{O}\!\left((s_i - s_j)^{3}\right)
    \;\geq\; 0,
\end{equation}
with equality if and only if $s_i = s_j$. Summed over $r$ merge pairs, the cumulative gap is non-negative and grows with $r$ as progressively more salience-asymmetric pairs enter the merge set.
\end{proposition}

\subsection{Computational Complexity}
The salience computation requires forming the affinity matrix $\mathbf{A} = \mathbf{X}\mathbf{X}^\top$, which costs $\mathcal{O}(N^2 d)$---the same order as self-attention. However, this cost is incurred \emph{once} per layer prior to merging and is immediately amortized: by reducing the sequence length from $N$ to $N - r_l$, the subsequent self-attention cost drops from $\mathcal{O}(N^2 d)$ to $\mathcal{O}((N-r_l)^2 d)$, yielding a net reduction that grows with $r_l$. \method{} thus maintains the plug-and-play efficiency of \tome{} with bounded overhead, as analyzed empirically in \S\ref{sec:throughput}.
\section{Experiments}
\label{sec:experiments}

\subsection{Main Results}
\label{sec:main_results}

To evaluate the efficacy of \method{}, we conduct FLOPs-matched comparisons against three representative training-free baselines: (i) \tome{}~\citep{bolya2023tome} as the pioneering framework; (ii) \textbf{DSM}~\citep{heo2024dsm}, representing depth-wise heuristic schedules; and (iii) \textbf{PiToMe}~\citep{tran2024pitome}, a recent energy-based importance-aware method. As reported in Table~\ref{tab:sota_comparison}, \method{} consistently outperforms all baselines across all six operating points, extending the observed accuracy--FLOPs Pareto frontier.

\begin{table}[h]
\centering
\caption{Comprehensive FLOPs-matched comparison across six tiers.
         \method{} results are mean $\pm$ std over 5 runs.$^{\ast}$
         \textbf{Bold} denotes the best accuracy at each tier.
         For DSM, $d$ denotes \texttt{delay\_layer}.}
\label{tab:sota_comparison}
\small
\makebox[\linewidth][c]{
\begin{tabular}{llccccc}
\toprule
FLOPs Tier & Method & Top-1 Acc & $\Delta$ Acc & FLOPs (G) & FLOPs $\downarrow$ & Speedup \\
\midrule
\multirow{4}{*}{${\sim}15.9$G}
    & \method{} ($r_{\max}{=}9$)   & \textbf{$84.70_{\pm 0.01}$\%} & \textbf{$-0.49$\%} & 15.91 &  8.8\% & $1.01\times$ \\
    & \tome{} ($r{=}3$)             & 84.69\%               & $-0.50$\% & 15.93 &  8.7\% & $1.10\times$ \\
    & PiToMe ($r{=}3$)              & 84.57\%               & $-0.62$\% & 15.93 &  8.7\% & $1.04\times$ \\
    & DSM ($r{=}12,\,d{=}6$)        & 84.42\%               & $-0.77$\% & 15.95 &  8.6\% & $1.06\times$ \\
\midrule
\multirow{4}{*}{${\sim}15.5$G}
    & \method{} ($r_{\max}{=}11$)  & \textbf{$84.62_{\pm 0.00}$\%} & \textbf{$-0.57$\%} & 15.54 & 10.9\% & $1.05\times$ \\
    & \tome{} ($r{=}4$)             & 84.44\%               & $-0.75$\% & 15.43 & 11.6\% & $1.10\times$ \\
    & PiToMe ($r{=}4$)              & 84.44\%               & $-0.75$\% & 15.43 & 11.6\% & $1.04\times$ \\
    & DSM ($r{=}18,\,d{=}6$)        & 83.68\%               & $-1.51$\% & 15.48 & 11.3\% & $1.09\times$ \\
\midrule
\multirow{4}{*}{${\sim}14.9$G}
    & \method{} ($r_{\max}{=}14$)  & \textbf{$84.49_{\pm 0.03}$\%} & \textbf{$-0.70$\%} & 14.98 & 14.1\% & $1.09\times$ \\
    & \tome{} ($r{=}5$)             & 84.12\%               & $-1.07$\% & 14.93 & 14.4\% & $1.15\times$ \\
    & PiToMe ($r{=}5$)              & 84.29\%               & $-0.90$\% & 14.93 & 14.4\% & $1.08\times$ \\
    & DSM ($r{=}16,\,d{=}5$)        & 83.35\%               & $-1.84$\% & 14.90 & 14.6\% & $1.12\times$ \\
\midrule
\multirow{4}{*}{${\sim}14.4$G}
    & \method{} ($r_{\max}{=}17$)  & \textbf{$84.28_{\pm 0.02}$\%} & \textbf{$-0.91$\%} & 14.39 & 17.5\% & $1.13\times$ \\
    & \tome{} ($r{=}6$)             & 83.67\%               & $-1.52$\% & 14.43 & 17.3\% & $1.19\times$ \\
    & PiToMe ($r{=}6$)              & 84.22\%               & $-0.97$\% & 14.43 & 17.3\% & $1.12\times$ \\
    & DSM ($r{=}14,\,d{=}4$)        & 82.51\%               & $-2.68$\% & 14.37 & 17.6\% & $1.21\times$ \\
\midrule
\multirow{4}{*}{${\sim}13.9$G}
    & \method{} ($r_{\max}{=}20$)  & \textbf{$84.25_{\pm 0.03}$\%} & \textbf{$-0.94$\%} & 13.94 & 20.1\% & $1.17\times$ \\
    & \tome{} ($r{=}7$)             & 83.11\%               & $-2.08$\% & 13.94 & 20.1\% & $1.25\times$ \\
    & PiToMe ($r{=}7$)              & 83.65\%               & $-1.54$\% & 13.94 & 20.1\% & $1.17\times$ \\
    & DSM ($r{=}17,\,d{=}4$)        & 81.05\%               & $-4.14$\% & 13.90 & 20.3\% & $1.26\times$ \\
\midrule
\multirow{4}{*}{${\sim}13.4$G}
    & \method{} ($r_{\max}{=}23$)  & \textbf{$84.13_{\pm 0.02}$\%} & \textbf{$-1.06$\%} & 13.42 & 23.1\% & $1.22\times$ \\
    & \tome{} ($r{=}8$)             & 82.46\%               & $-2.73$\% & 13.44 & 23.0\% & $1.31\times$ \\
    & PiToMe ($r{=}8$)              & 83.74\%               & $-1.45$\% & 13.44 & 23.0\% & $1.23\times$ \\
    & DSM ($r{=}20,\,d{=}4$)        & 80.57\%               & $-4.62$\% & 13.47 & 22.8\% & $1.28\times$ \\
\bottomrule
\end{tabular}
}
\vspace{4pt}
\begin{minipage}{\linewidth}
\small
$^{\ast}$\,The ImageNet-1k validation set is fully deterministic; randomness arises solely from the inference computation. Specifically, standard deviation reflects variance from (i)~stochastic tie-breaking in the adaptive-$r$ decision when similarity scores are equal, and (ii)~non-deterministic CUDA floating-point operations during the forward pass.
\end{minipage}
\end{table}

\paragraph{Monotonic gap widening.}
The accuracy advantage of \method{} over \tome{} \emph{widens monotonically} as compression intensifies (Table~\ref{tab:gap_comparison}). This trend is consistent with Proposition~\ref{prop:reconstruction}: salience-asymmetric token pairs incur higher reconstruction error under uniform averaging, and this gap grows as larger $r$ forces progressively more dissimilar pairs into the merge set.
 
\begin{table}[h]
\centering
\caption{Accuracy gap (\method{} $-$ \tome{}) at matched FLOPs tiers.}
\label{tab:gap_comparison}
\small
\begin{tabular}{lcccccc}
\toprule
FLOPs Tier
    & ${\sim}15.9$G & ${\sim}15.5$G & ${\sim}14.9$G
    & ${\sim}14.4$G & ${\sim}13.9$G & ${\sim}13.4$G \\
\midrule
Gap (\method{} $-$ \tome{})
    & $+0.01$\%p & $+0.18$\%p & $+0.37$\%p
    & $+0.61$\%p & $+1.14$\%p & $+1.67$\%p \\
\bottomrule
\end{tabular}
\end{table}

\paragraph{DSM failure mode.}
DSM's rigid, delayed merging schedule incurs a substantial $-4.62$\%p degradation at ${\sim}13.4$G in aggressive regimes. 
In contrast, \method{}'s adaptive intensity limits the degradation to $-1.06$\%, highlighting the benefit of input- and layer-adaptive reduction counts. 
 
\begin{remark}[Strong Empirical Performance]
\method{} outperforms all baselines at matched FLOPs in the aggressive-compression regime, extending the accuracy--FLOPs Pareto frontier precisely where existing methods degrade most sharply.
\end{remark}
 
\subsection{Throughput--Accuracy Trade-off}
\label{sec:throughput}
 
\begin{figure}[t]
\centering
\includegraphics[width=\linewidth]{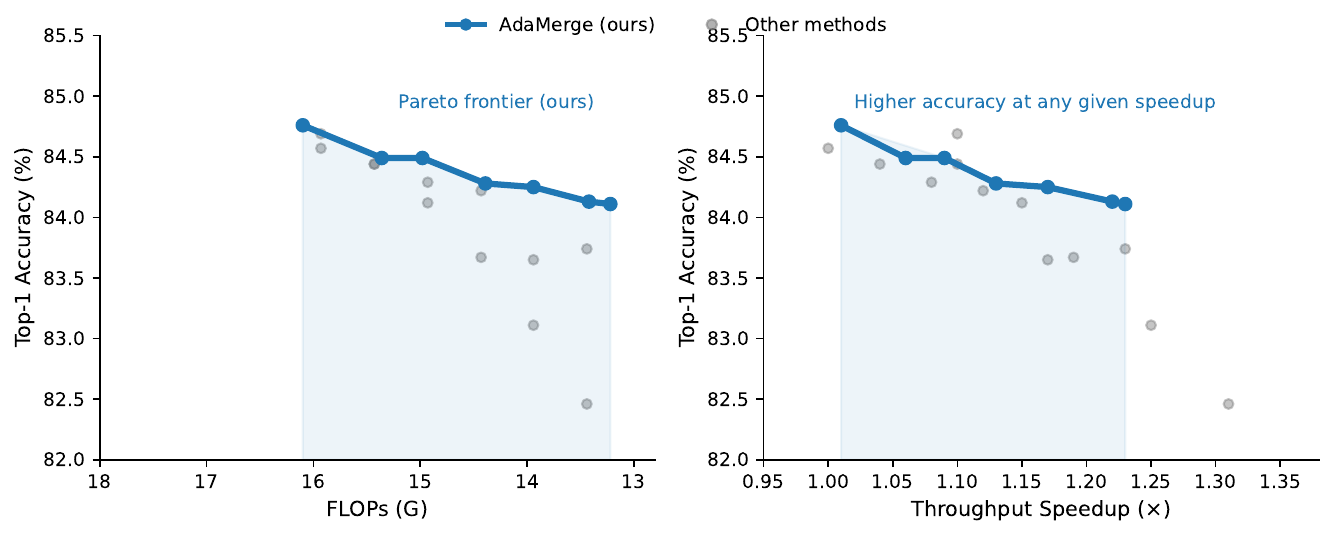}
\caption{
    \textbf{FLOPs--Accuracy and Speedup--Accuracy trade-off curves.}
    \emph{Left}: \method{} consistently extends the FLOPs--accuracy Pareto
    frontier above all baselines.
    \emph{Right}: \method{} and \tome{} occupy complementary regions of the
    speedup--accuracy frontier. \method{} offers higher accuracy at any given
    speedup level; \tome{} achieves higher speedup at the cost of accuracy.
}
\label{fig:pareto}
\end{figure}

Figure~\ref{fig:pareto} presents the FLOPs--accuracy and speedup--accuracy trade-off curves for all methods.
The two panels reveal complementary strengths.
 
\paragraph{FLOPs--Accuracy (Figure~\ref{fig:pareto} Left).}
\method{} consistently occupies the upper envelope of the FLOPs--accuracy frontier across all operating points.
At 23\% FLOPs reduction, \tome{}'s accuracy floor is 82.46\%, while \method{} sustains 84.13\%---a gap of 1.67\%p.
This gap widens monotonically with compression, suggesting that ADAMERGE provides a more favorable accuracy--FLOPs trade-off in high-compression regimes.
 
\paragraph{Speedup--Accuracy (Figure~\ref{fig:pareto} Right).}
The throughput picture is deliberately honest: \method{} and \tome{} occupy \emph{complementary} regions of the speedup--accuracy frontier.
At every speedup level achievable by \tome{}, \method{} offers a higher-accuracy alternative at slightly reduced throughput ($5$--$9$\%).
Conversely, if throughput is the primary constraint, \tome{} achieves up to $1.31\times$ speedup at the cost of substantially lower accuracy.
The choice between methods is therefore application-dependent: latency-critical deployments may prefer \tome{}'s tighter speedup, while accuracy-critical deployments (e.g., medical imaging, ADAS) benefit from \method{}'s preservation of discriminative information.

\subsection{Layer-wise Merging Distribution}
\label{sec:layer_dist}
 
\method{}'s emergent merging distribution exhibits three qualitatively
distinct regimes depending on $r_{\max}$.
 
\textbf{Regime 1 (mild compression, $r_{\max} \leq 11$).}
The final block receives the largest merging allocation
($\mathbb{E}[r_{11}] = 5.18$ at $r_{\max}{=}9$),
while shallow layers merge conservatively.
This suggests that, under a limited budget, the adaptive mechanism tends to assign more merges
to deeper layers, where token representations may contain higher redundancy after repeated
self-attention operations.
 
\textbf{Regime 2 (moderate compression, $r_{\max} \in \{14, 17\}$).}
The distribution inverts: blocks 0--1 become the most active
($\mathbb{E}[r_1] = 4.71$--$5.87$) and block 11 contracts sharply
($\mathbb{E}[r_{11}] = 3.12$--$0.84$),
consistent with the inductive bias that shallow layers
possess high spatial redundancy.
 
\textbf{Regime 3 (aggressive compression, $r_{\max} \geq 20$).}
Layers $l \geq 10$ become completely inactive
($\mathbb{E}[r_l] = 0$): aggressive shallow merging exhausts the
token budget before reaching the deepest blocks.
 
\textbf{Cross-regime invariant.}
Block $l=2$ records the lowest $\mathbb{E}[r_l]$ among active layers 
across all six settings, indicating that layer 2 consistently produces the most
diverse intermediate representations regardless of compression level.
 
\begin{remark}[Interpretation]
The three-regime structure is an emergent property of the
sigmoid-based adaptive-$r$ schedule interacting with the model's
layer-wise redundancy profile---no manual per-layer budget was
specified. This self-organizing behaviour provides finer-grained
allocation than hand-tuned schedules~(e.g., DSM's fixed
\texttt{delay\_layer}) and aligns with known patterns of attention
sparsification in deep models~\citep{caron2021dino}.
This input-adaptive behaviour is further corroborated qualitatively:
\method{} compresses up to 72.4\% of tokens for a uniform
polar-bear image while retaining dense structure in complex
multi-object scenes.
\end{remark}
\section{Ablation Studies}
\label{sec:ablation}
 
\subsection{Component-Wise Decomposition}
We isolate the contributions of salience weighting (SW) and adaptive
$r$ (Adp) via a component ablation at FLOPs $\approx$15.0G
(Table~\ref{tab:ablation_component}).
Salience weighting alone provides matching guidance but no compression
budget control; adaptive $r$ alone provides budget control but lacks
information-preserving aggregation.
The combination is essential not because the components are additive,
but because adaptive $r$ without salience guidance allocates the merge
budget to suboptimal pairs ($-0.48$\%p vs.\ \tome{}), while salience
without adaptive $r$ cannot exploit the per-image redundancy structure
($-0.15$\%p vs.\ \tome{}).

\begin{table}[h]
\centering
\caption{Component ablation ($r_{\max}{=}14$, \tome{} $r{=}5$,
         FLOPs $\approx$15.0G).
         \textbf{Bold} denotes the best accuracy.
         $^\ddagger$Adaptive-$r$ variants merge a variable number of tokens;
         Full \method{} at $r_{\max}{=}24$ is added to match the average
         merge count of \emph{Adaptive r only} ($\approx$69 tokens),
         isolating the effect of merging quality from merging quantity.}
\label{tab:ablation_component}
\small
\begin{tabular}{lcccccc}
\toprule
Configuration & SW & Adp & Top-1 Acc & $\Delta$ vs.\ \tome{} & Avg.\ Merges \\
\midrule
\tome{} ($r{=}5$)            & \xmark & \xmark & 84.10\%          & --          & 60.0 \\
$+$ Salience only ($r{=}5$)  & \cmark & \xmark & 83.95\%          & $-0.15$\%p  & 60.0 \\
$+$ Adaptive $r$ only$^\ddagger$
                              & \xmark & \cmark & 83.62\%          & $-0.48$\%p  & 68.8 \\
\midrule
\textbf{Full \method{}} ($r_{\max}{=}14$)$^\ddagger$
                              & \cmark & \cmark & \textbf{84.43\%} & $+0.33$\%p  & 47.2 \\
Full \method{} ($r_{\max}{=}24$)$^\ddagger$
                              & \cmark & \cmark & 84.13\%          & $+0.03$\%p  & 70.7 \\
\bottomrule
\end{tabular}
\begin{minipage}{\linewidth}
\vspace{4pt}
\raggedright\footnotesize
$^\ddagger$ At comparable merge counts, Full \method{} ($r_{\max}{=}24$,
70.7 tokens merged, 84.13\%) outperforms \emph{Adaptive r only}
(68.8 tokens merged, 83.62\%) by $+0.51$\%p,
even while merging \emph{more} tokens,
confirming that the accuracy gain stems from the quality of merging
decisions rather than from a reduced merge count.
\end{minipage}
\end{table}

\paragraph{Mechanistic coupling.}
The Adp-only underperformance ($-0.48$\%p) reflects a coherence
requirement: the redundancy proxy $\bar{S}_l$ is a reliable signal only
when the underlying similarities reflect information content rather than
raw feature proximity.
Without salience weighting, cosine similarity conflates
\emph{similar-but-informative} pairs with \emph{similar-and-redundant}
ones, causing adaptive-$r$ to over-merge salience-asymmetric layers.
The two components are therefore \emph{coupled by construction}:
salience weighting supplies the signal quality that adaptive-$r$ needs
to allocate its budget meaningfully, and adaptive-$r$ supplies the
per-image schedule that salience weighting needs to avoid uniform
over-compression.

To rule out the hypothesis that \method{}'s gain stems merely from
merging fewer tokens, we compare at comparable merge counts:
Full \method{} at $r_{\max}{=}24$ (70.7 tokens merged, 84.13\%)
outperforms \emph{Adaptive $r$ only} (68.8 tokens merged, 83.62\%)
by $+0.51$\%p \emph{despite merging more tokens}, confirming that
the advantage arises from merging quality rather than reduced count.
 
\subsection{Persistence under Fine-Tuning}
\label{subsec:finetuning_analysis}
Fine-tuning the backbone jointly with the merging module (30 epochs,
AdamW, LR $= 5 \times 10^{-6}$, cosine decay) confirms
that \method{}'s advantage is structural, not an artefact of the
training-free regime.
The accuracy gap over \tome{} \textbf{widens monotonically} with
compression intensity ($+0.42 \to +0.57 \to +0.72$\%p at
${\sim}15.4$G, ${\sim}15.0$G, ${\sim}14.4$G respectively),
mirroring the training-free pattern.

\subsection{Efficiency and Stability of the Refinement Protocol}
\label{subsec:refinement_analysis}
The iterative refinement protocol enforces self-consistency between
precomputed statistics and the actual inference-time token distribution.
Two passes from the default initialisation are sufficient: the first pass
establishes a coarse estimate of the layer-wise redundancy distribution,
and the second pass converges to a stable fixed point.
A third pass overshoots the fixed point and induces oscillation
($\sim\!1$\%p accuracy drop), so two passes is the recommended protocol.
A 1\% calibration subset is sufficient: varying the subset from 0.5\%
to 5\% changes Top-1 accuracy by at most $0.09$\%p ($84.21$--$84.30$\%)
and produces nearly identical per-layer statistics
($\Delta\mu_l < 0.001$ across all layers),
confirming robustness to sampling variance.
Shallow-layer statistics remain stable across $r_{\max}$ settings
while deep-layer statistics self-calibrate to the compression regime.
 
\section{Conclusion}
\label{sec:conclusion}
We introduced \method{}, a training-free token-merging framework that
addresses two primary limitations of prior merging methods: their
insensitivity to token importance and their reliance on fixed
compression schedules. By weighting the bipartite matching score with
feature-affinity salience and dynamically modulating the per-layer
merge count via pre-computed redundancy statistics, \method{} preserves
high-salience information while adapting compression to both the input
image and the current layer's semantic density. We further provide a
theoretical grounding: under an isotropic noise model,
salience-proportional aggregation yields a non-negative reduction in
feature-reconstruction error relative to uniform averaging, with the
gap growing as compression intensifies
(Proposition~\ref{prop:reconstruction}).

On ImageNet-1k with ViT-B/16, \method{} consistently extends the
accuracy--FLOPs Pareto frontier over \tome{}, PiToMe, and DSM across
all evaluated operating points. The advantage widens monotonically with
compression: at the ${\sim}13.4$G FLOPs tier, \method{} sustains a
Top-1 degradation of only $-1.06\%$, compared to $-2.73\%$ for
\tome{} and $-4.62\%$ for DSM. This monotonic widening---reproduced
under both training-free and fine-tuned settings---is consistent with
the hypothesis that the benefit arises from the structural coupling of
salience-aware matching and adaptive-$r$ allocation, rather than from
incidental reduction in merge count.

More broadly, these results suggest that the token-equality assumption
embedded in current merging pipelines is a source of accuracy loss
that becomes increasingly costly as compression grows. \method{}
demonstrates that relaxing this assumption in a strictly training-free
regime is practically feasible under a theoretically motivated
framework, offering a drop-in replacement for existing merging
schedules without requiring retraining or architectural modification.

The current evaluation focuses on ImageNet-1k classification with
ViT-B/16; while the salience formulation and adaptive-$r$ mechanism
are architecture-agnostic by construction, their behaviour on larger
backbones (ViT-L), self-supervised checkpoints (DINOv2, MAE), and
dense prediction tasks (segmentation, detection) remains to be
empirically verified. Key open problems include reducing the 5--9\%
throughput overhead relative to \tome{}---which stems from salience
computation and per-layer adaptive branching and could be mitigated
through fused-kernel implementations or learnable layer-wise
$r_{\max}$ schedules---extending \method{} to video and multi-modal
Transformers where token redundancy structure differs substantially
from the spatial-only setting, and jointly optimizing the per-layer
$r_{\max}$ schedule through fine-tuning to more fully exploit the
reduced-token regime.


{
\small
\bibliography{reference}





\end{document}